# $L_2$-optimal image interpolation and its applications to medical imaging.

Oleg S. Pianykh, PhD

**Abstract:**
Digital medical images are always displayed scaled to fit particular view. Interpolation is responsible for this scaling, and if not done properly, can significantly degrade diagnostic image quality. However, theoretically-optimal interpolation algorithms may also be the most time-consuming and impractical. We propose a new approach, adapted to the needs of digital medical imaging, to combine high interpolation speed and superior $L_2$-optimal image quality.

PART I: $L_2$-OPTIMAL KERNELS

I. INTRODUCTION

The choice of digital image interpolation is nothing but the choice of the interpolation optimality criteria (interpolation kernel smoothness, order of approximation, kernel size and complexity – to name a few). This has been demonstrated by many excellent studies [Lehmann], [Thévenaz], [Blu], and lead to a series of widely accepted interpolation algorithms. The goal of our analysis was to review interpolation optimality in the light of digital radiology, suggesting the best practical solution.

Interpolation theory is driven by the study of the interpolation kernel functions: pixel values in the interpolated image $J(x,y)$ are computed as the convolution of the original pixels $I(x,y)$ with continuous 2D interpolation kernel function $h_{2D}(x,y)$:

$$J(x,y) = \sum_k \sum_n I(k,n) \times h_{2D}(x-k, y-n)$$

For computational simplicity and isotropy, symmetric separable kernels are preferred,

$$\left. \begin{array}{l} h_{2D}(x,y) = h(x) \times h(y), \; h(x)=h(-x), \\ \text{with finite support}^1 \; L < \infty : \\ h(x) = 0 \quad \forall x : |x| > L \end{array} \right\} \quad \text{(Eq.C1)}$$

This results in

$$J(x,y) = \sum_{k=0}^{L} h(x-k) \left[ \sum_{n=0}^{L} I(k,n) \times h(y-n) \right] \quad \text{(Eq.I)}$$

In the most basic case, we want (Eq.I) to preserve $I(x,y)$ average and be identity for the overlapping pixel values, which leads to the following fundamental interpolating kernel conditions [Lehmann]:

---

[1] In imaging applications support $L$ is commonly chosen as 1, 2 or 3, rarely 4; $L=2$ and $L=3$ being the most typical choices in medical imaging software.

$$h(n) = \delta_n^0,$$

$$\sum_{k=-L}^{L} h(x+k) = 1, \quad 0 \le x \le 1, \quad n,k \in Z, \quad L \in N \tag{Eq.C2}$$

Given conditions (Eq.C1) and (Eq.C2), the most common approach to selecting $h(x)$ is based on preserving the local frequency spectrum at each pixel $(k,n)$. This suggests Fourier analysis, and Fourier transform of the sinc kernel $h_s(x) = \sin(\pi x)/(\pi x) = Sinc(x)$ produces the ideal rectangular ("box") frequency response

$$\Pi(t) = \begin{cases} 1, & t < 1/2 \\ 1/2, & t = 1/2 \\ 0, & t > 1/2 \end{cases}$$

$Sinc(x)$ satisfies (Eq.C2), but unfortunately has infinite support $L$, and cannot be used practically. Therefore, interpolation kernel research concentrated on building non-$Sinc$ kernels $h(x)$ with other optimal properties (smoothness, order of approximation, proximity to $L$-truncated $Sinc$), achievable on finite support. This approach has proven to be very fruitful, generating a wealth of kernel functions, designs, and optimality criteria [Lehmann]. With little exceptions, all those $h(x)$ were sought in the space of piecewise polynomials, leading to such well-known interpolation kernels as Keys, B-splines, MOMS[2] [Keys], [Lehmann], [Thévenaz], [Blu], [Blu2], [Meijering], [Shi]. In particular, the following interpolation kernels[3] have become very popular:

Linear: $h_{linear}(x) = 1 - x, \quad 0 \le x \le 1, \quad L = 1$

Keys: $h_{Keys}(x) = \begin{cases} (a+2)x^3 - (a+3)x^2 + 1, & 0 \le x \le 1 \\ ax^3 - 5ax^2 + 8ax - 4, & 1 \le x \le 2 \end{cases} \quad a = \tfrac{1}{2}, \quad L = 2$ (Eq.Popular)

Cubic6: $h_{Cubic6}(x) = \dfrac{1}{5}\begin{cases} 6x^3 - 11x^2 + 5, & 0 \le x \le 1 \\ -3x^3 + 16x^2 - 27x + 14, & 1 \le x \le 2 \\ x^3 - 8x^2 + 21x - 18, & 2 \le x \le 3 \end{cases} \quad L = 3$

However, the practicalities of digital radiology may impose their own constraints on the interpolation criteria selection, as we will investigate in our work. With this in mind, we subdivided our analysis into two principal parts: developing the most general (unconstrained) $L_2$-optimal interpolation kernels, and studying their application to the medical imaging.

## II. $L_2$-OPTIMAL INTERPOLATION KERNELS

We introduce $L_2$-optimal interpolation kernels as theoretically-optimal way of preserving image frequency content with linear interpolation (Eq.I). For positive integer $L$, let $\Lambda_L$ be the set of all symmetric functions with finite support $L$, satisfying (Eq.C2). For $\forall h(x) \in \Lambda_L$ and its Fourier transform $F_h(t) = \int_{-\infty}^{\infty} h(x) e^{-2\pi i x t} dx$ we define *frequency approximation error* (FAE) function as $E(h) = \left( \int_{t=-\infty}^{+\infty} (F_h(t) - \Pi(t))^2 dt \right)^{1/2}$. We define

$L_2$-optimal interpolation kernel $H_L(x) \in \Lambda_L$ as

$$H_L(x) = \arg\min_{h(x) \in \Lambda_L} E(h). \tag{Eq.L2Opt}$$

FAE function $E(h)$ measures the accuracy of $h(x)$ in the frequency domain. By definition $E(Sinc)=0$ (ideal interpolation with no frequency loss), and $E(0)=1$ ($h(x)=0$, "no interpolation" case, full frequency loss).

---

[2] MOMS kernels relax interpolation conditions (Eq.C2) to provide optimal order of interpolation.
[3] For notational simplicity we provide only non-zero segments of $h(x)$.

Therefore in terms of the kernel support size *L* this can be rephrased as $E(L=\infty)=0$ and $E(L=0)=1$. We would like to find how $E(L)$ depends on the finite support size *L*.

*Theorem 1 (L$_2$-optimal interpolation kernel):*
For any support size *L*, the optimal kernel $H_L(x)$ in (Eq.L2Opt) exists and is uniquely defined by the following function:

$$H_L(x) = \begin{cases} Sinc(x) + \dfrac{1}{2L}\left[1 - \sum_{k=0}^{2L-1} Sinc\left((-1)^{k+n} x + \left\lfloor\dfrac{k+1}{2}\right\rfloor + (-1)^{k+n+1}\left\lfloor\dfrac{n+1}{2}\right\rfloor\right)\right] \\ n/2 \le x < (n+1)/2, \quad n = 0 \ldots 2L-1 \end{cases}$$ (Eq.HL)

**Proof**
From the orthogonality of Fourier transform *F* (Parseval's theorem) and $F(Sinc(x))=\Pi(t)$, taking into account finite support and symmetry of *h(x)*, we rewrite

$$E^2(h) = \int_{t=-\infty}^{+\infty}(F_h(t)-\Pi(t))^2 dt = \int_{x=-\infty}^{\infty}(h(x)-Sinc(x))^2 dx$$

$$= 2\int_0^L (h(x)-Sinc(x))^2 dx + 2\int_L^{+\infty} Sinc^2(x) dx = 2(E_1(h)+E_2)$$ (Eq.FAE)

The second term $E_2$ does not depend on *h(x)*. The first term can be rewritten as

$$E_1(h) = \int_0^L (h(x)-Sinc(x))^2 dx = \sum_{k=0}^{2L-1}\left[\int_{k/2}^{(k+1)/2}(h(x)-Sinc(x))^2 dx\right]$$ (Eq.E1)

On each half-unit segment [*k*/2, (*k*+1)/2] we change the integration variable as $x' = (-1)^k\left(x - \left\lfloor\dfrac{k+1}{2}\right\rfloor\right)$, to shift all integration segments to [0, ½]:

$$E_1(h) = \sum_{k=0}^{2L-1}\left[\int_{k/2}^{(k+1)/2}(h(x)-Sinc(x))^2 dx\right] =$$

$$= \sum_{k=0}^{2L-1}\left[\int_0^{1/2}\left(h\left(\left\lfloor\dfrac{k+1}{2}\right\rfloor+(-1)^k x\right) - Sinc\left(\left\lfloor\dfrac{k+1}{2}\right\rfloor+(-1)^k x\right)\right)^2 dx\right]$$ (Eq.E1b)

$$= \int_0^{1/2}\left[\sum_{k=0}^{2L-1}\left(h\left(\left\lfloor\dfrac{k+1}{2}\right\rfloor+(-1)^k x\right) - Sinc\left(\left\lfloor\dfrac{k+1}{2}\right\rfloor+(-1)^k x\right)\right)^2\right] dx$$

At the same time from (Eq.C2), finite support, and symmetry of *h(x)*:

$$S(x) = \sum_{k=-L}^{L} h(x+k) = \sum_{k=0}^{2L-1} h\left(\left\lfloor\dfrac{k+1}{2}\right\rfloor+(-1)^k x\right) = 1, \quad 0 \le x \le 1.$$ (Eq.E1c)

Therefore let's introduce functions

$$h_k(x) = \begin{cases} h\left(\left\lfloor\dfrac{k+1}{2}\right\rfloor+(-1)^k x\right), & x \in [k/2, (k+1)/2] \\ 0, & \text{otherwise} \end{cases}$$ (Eq.hk)

and

$$s_k(x) = Sinc\left(\left\lfloor\dfrac{k+1}{2}\right\rfloor+(-1)^k x\right).$$ (Eq.sk)

Then each $h_k(x)$ uniquely defines *h(x)* on [*k*/2, (*k*+1)/2], and

$$h(x) = \sum_{k=0}^{2L-1} h_k \left( (-1)^k \left( x - \left\lfloor \frac{k+1}{2} \right\rfloor \right) \right)$$

From (Eq.E1c) $\sum_{k=0}^{2L-1} h_k(x) = 1$, or $h_{2L-1}(x) = 1 - \sum_{k=0}^{2L-2} h_k(x)$, and substituting $h_{2L-1}(x)$ in (Eq.E1b) yields:

$$E_1(h) = \int_0^{1/2} \left[ \sum_{k=0}^{2L-2} (h_k(x) - s_k(x))^2 + \left( -1 + s_{2L-1}(x) + \sum_{k=0}^{2L-2} h_k(x) \right)^2 \right] dx \qquad \text{(Eq.var)}$$

This is a well-defined variance optimization problem for a set of (2L-1) independent functions $h_k(x)$, k=0…2L-2. The minimum of $E_1(h)$ in (Eq.var) should satisfy Euler's condition $\partial E_1 / \partial h_n = 0$ [Bronshtein], leading to the following system of linear equations:

$$\sum_{k=0}^{2L-2} h_k(x) + h_n(x) = 1 + s_n(x) - s_{2L-1}(x), \quad n=0\ldots2L\text{-}2. \qquad \text{(Eq.lin)}$$

If we define $U(x) = \sum_{k=0}^{2L-2} h_k(x)$ and $W(x) = \sum_{k=0}^{2L-2} s_k(x)$, then by adding all the equations in (Eq.lin) we find

$2LU(x) = (2L-1)(1 - s_{2L-1}(x)) + W(x)$, or

$U(x) = \dfrac{(2L-1)(1 - s_{2L-1}(x)) + W(x)}{2L}$, and

$$h_n(x) = 1 + s_n(x) - s_{2L-1}(x) - U(x) = 1 + s_n(x) - s_{2L-1}(x) - \frac{(2L-1)(1 - s_{2L-1}(x)) + W(x)}{2L} =$$

$$= \frac{1 - s_{2L-1}(x) + 2L s_n(x) - W(x)}{2L} = s_n(x) + \frac{1}{2L}\left(1 - \sum_{k=0}^{2L-1} s_k(x)\right) \qquad \text{(Eq.sol)}$$

Then $h_{2L-1}(x) = 1 - \sum_{k=0}^{2L-2} h_k(x) = s_{2L-1}(x) + \dfrac{1}{2L}\left(1 - \sum_{k=0}^{2L-1} s_k(x)\right)$, thus following the same equation (Eq.sol) for n=2L-1 From (Eq.sk)

$$h_n(x) = Sinc(x) + \frac{1}{2L}\left[1 - \sum_{k=0}^{2L-1} Sinc\left((-1)^{k+n} x + \left\lfloor \frac{k+1}{2} \right\rfloor + (-1)^{k+n+1}\left\lfloor \frac{n+1}{2} \right\rfloor\right)\right] =$$

$= Sinc(x) + T_n(x)$

∎

As one can see, on each segment [n/2, (n+1)/2], optimal $H_L(x)$ consists of the "ideal kernel" $Sinc(x)$ and the "aliasing" term $T_n(x)$ (penalizing $H_L(x)$ for the finite support L). Moreover, one can easily show that $T_n(x)$ terms can be gathered in the following continuous function

$T_L(x) = H_L(x) - Sinc(x) =$

$$= \begin{cases} T_n(x), \\ n/2 \le x < (n+1)/2, \ n = 0\ldots 2L-1 \end{cases} = \qquad \text{(Eq.TL)}$$

$$= \begin{cases} \dfrac{1}{2L}\left[1 - \sum_{k=0}^{2L-1} Sinc\left((-1)^{k+n} x + \left\lfloor \frac{k+1}{2} \right\rfloor + (-1)^{k+n+1}\left\lfloor \frac{n+1}{2} \right\rfloor\right)\right] \\ \qquad n/2 \le x < (n+1)/2, \quad n = 0\ldots 2L-1 \end{cases}$$

vanishing to 0 as L grows (see Figure 1, left).

Knowing the exact formula for the least-square-optimal kernel enables us to study those kernels more closely. In particular, we can find the minimal frequency approximation error as a function of the kernel support size $L$.

*Corollary 1:*
Minimal frequency approximation error for finite support $L$ is given by (Figure 1, right):

$$E_L = E(H_L) = \sqrt{2\sum_{n=0}^{2L-1}\left[\int_{n/2}^{(n+1)/2} T_n^2(x)dx\right] + 2\int_L^{+\infty} Sinc^2(x)dx} \qquad \text{(Eq.EL)}$$

**Proof**
By substituting the optimal $h(x)=H_L(x)$ from (Eq.HL) into (Eq.FAE). Note that "no interpolation" case for $L=0$ can be also defined as $E_0 = E(0) = \sqrt{2\int_0^{+\infty} Sinc^2(x)dx} = \sqrt{2\int_0^{+\infty}\Pi^2(t)dt} = 1$, as we described previously.

∎

The practical meaning of FAE $E_L=E(H_L(x))$ is straightforward: any $L$-supported interpolation kernel $h(x)$ can reproduce at most $1-E_L$ of the image frequency content (measured in $L_2$ norm). Therefore we believe that $E_L(L)$ gives better measure of kernel quality than smoothness or boundary conditions (such as maximum order) – it simply reflects how close we get to the ideal $\Pi(t)$ filter in the least-square sense.

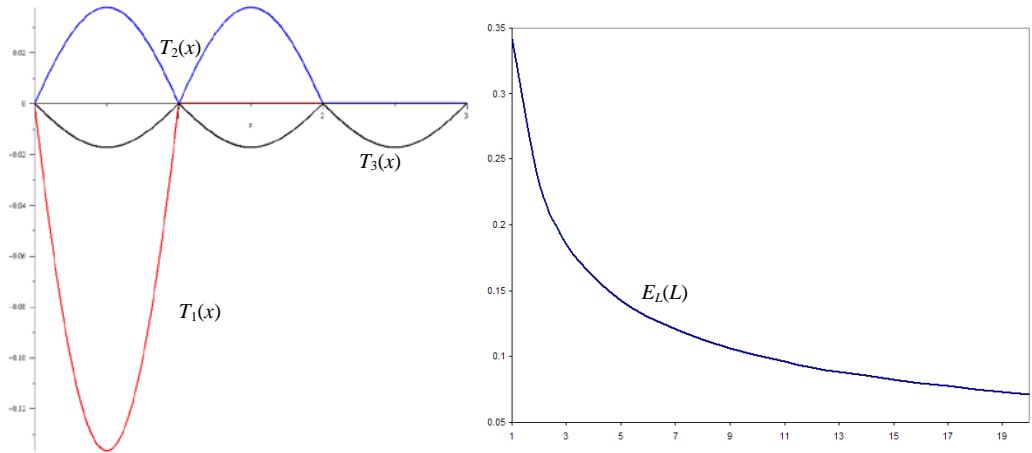

**Figure 1: Left: $T_L(x)=H_L(x)-Sinc(x)$ for $L=1$ (red), $L=2$ (blue), and $L=3$ (black). Right: Optimal error $E_L$ for $L=1\ldots20$.**

For practical applications, we have found the following approximate formula for (Eq.EL):
$$\hat{E}_L = 0.335L^{-0.5258}, \qquad \text{(Eq.ELap)}$$
which for $L\leq 15$ deviates from the true (Eq.EL) by less than 2%.

As one can see on Figure 1(right), the most significant reduction in $E_L(L)$ occurs for small $L$. Moreover, we are most interested in small $L$ to minimize the computational effort in (Eq.I). Therefore, we chose to compare our optimal kernels in (Eq.HL) with the popular ones in (Eq.Popular) for $L=1$, 2, and 3. From (Eq.HL) we derive

*Corollary 2:*
$L_2$-optimal interpolating kernels $H_1(x)$, $H_2(x)$, and $H_3(x)$ (for support $L=1$, 2, and 3 respectively) are
$$H_1(x) = \tfrac{1}{2}(1 + Sinc(x) - Sinc(1-x)), \quad 0 \leq x \leq 1$$
$$H_2(x) = \tfrac{1}{4}\begin{cases} 1 + 3Sinc(x) - Sinc(1-x) - Sinc(1+x) - Sinc(2-x), & 0 \leq x \leq 1 \\ 1 + 3Sinc(x) - Sinc(1-x) - Sinc(2-x) - Sinc(3-x), & 1 \leq x \leq 2 \end{cases}$$

$$H_3(x) = \frac{1}{6} \begin{cases} 1 + 5\,Sinc(x) - Sinc(1-x) - Sinc(1+x) - Sinc(2-x) - Sinc(2+x) - Sinc(3-x), & 0 \le x \le 1 \\ 1 + 5\,Sinc(x) - Sinc(1-x) - Sinc(1+x) - Sinc(2-x) - Sinc(3-x) - Sinc(4-x), & 1 \le x \le 2 \\ 1 + 5\,Sinc(x) - Sinc(1-x) - Sinc(2-x) - Sinc(3-x) - Sinc(4-x) - Sinc(5-x), & 2 \le x \le 3 \end{cases} \qquad \text{(Eq.Cor2)}$$

These kernels are shown in Figure 2 (left), along with their Fourier transforms (right). As expected, all $H_k(x)$ satisfy (Eq.C2), and their Fourier transforms provide the best least-squares approximation to $\Pi(t)$ for the given support size $L$.

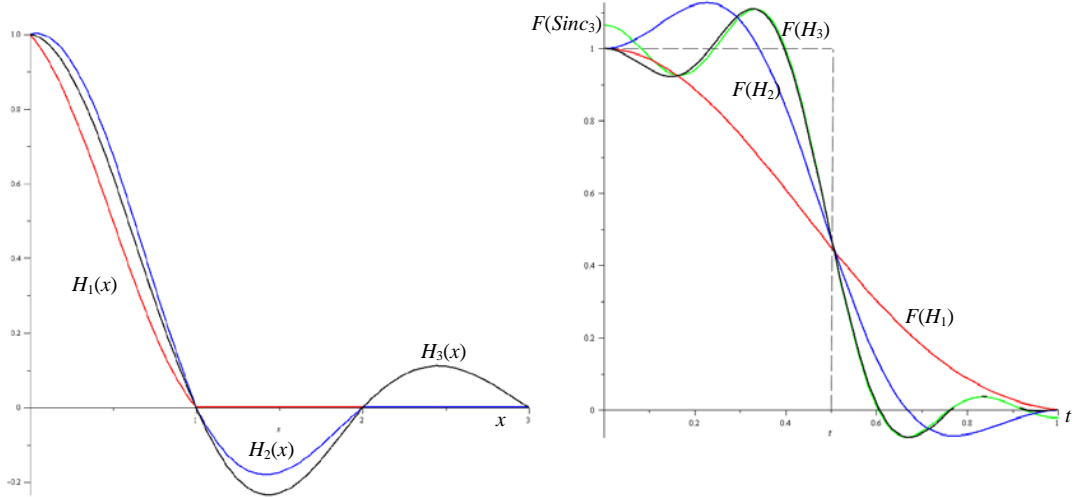

**Figure 2: Left: $L_2$-optimal kernels $H_1(x)$ (red), $H_2(x)$ (blue), and $H_3(x)$ (black) Right: Their Fourier transforms. Fourier transform graph also includes the Fourier transform for $Sinc_3(x)$ (green) - $Sinc(x)$, truncated to finite $L=3$ support.**

Note that $H_1(x)$ is very close to the linear interpolation kernel $h_{linear}(x)=1-x$, and $H_3(x)$ would be undistinguishable from $Sinc(x)$ on the left Figure 2 plot. However, their differences become more apparent in the frequency domain – compare $F(H_3)$ and $F(Sinc_3)$ for small $t$ on Figure 2 right. For this reason we compare the optimal $H_k(x)$ and popular (Eq.Popular) kernels in the Fourier domain, as illustrated in Figure 3.

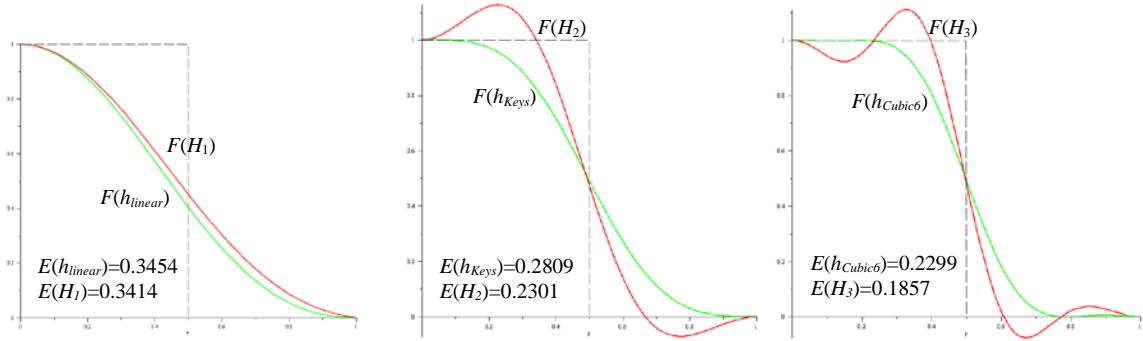

**Figure 3: Comparing $H_k(x)$ to popular interpolation kernels in Fourier domain: $H_1(x)$ vs. $h_{linear}(x)$ (L=1), $H_2(x)$ vs. $h_{Keys}(x)$ (L=2), $H_3(x)$ vs. $h_{Cubic6}(x)$ (L=3). Fourier transforms for $H_k(x)$ are shown in red, and FAE values are shows for each kernel.**

In particular, all Fourier transforms $F(H_k)$ for $k>1$ have local extrema, while traditional kernel design preference was to avoid them, forcing (nearly) monotone $F(h)$. However, we believe that Fourier monotonicity is a rather subjective choice: it will not make the $F(h)$ frequencies more equal, unless $F(h)$ is as flat as $\Pi(t)$. Therefore, while substantial peaks in $F(h)$ should be avoided, balancing them around constant $\Pi(t)$ segments can make more practical sense.

Another interesting observation can be derived from comparing the frequency approximation errors $E(h)$ for optimal (Eq.Cor2) and popular (Eq.Popular) kernels ($E$-values on Figure 3). Not only the optimal kernels produce smaller errors (as expected), but in some cases optimal kernels on smaller support ($H_2(x)$ for $L=2$, $E(H_2)=0.2301$) can preserve as much frequency content as best-known kernels on larger support ($h_{Cubic6}(x)$ on $L=3$, $E(h_{Cubic6})=0.2299$). This strongly speaks in favor of the optimal kernels: using smaller support sizes, they deliver faster interpolation, while preserving the same amount of frequency content.

PART II: APPLICATIONS IN RADIOLOGY

In this second part of our study, we review interpolation applications in digital radiology, and practical constraints arising from them.

I. INTERPOLATION FOR MEDICAL IMAGES

Pure research left apart, medical digital images are manipulated at the following three principal stages of the radiology workflow [Pianykh]:
1. Image acquisition
2. Quality control (QC)
3. Radiological viewing ( "reading")

Image acquisition is performed at the digital modalities (MR, CT, CR, …), and typically amounts to converting proprietary image formats to DICOM-compliant bitmaps. This conversion is definitely beyond the scope of our paper; besides, it may not use interpolation at all, deploying manufacturer or modality-specific transforms (such as Radon projection for CT).

QC is typically performed on technologists' workstations, right after the images are acquired, and before they are archived to the PACS[4] server. QC very rarely deals with any data resampling: if scanning protocols were set right, the probability of getting wrong images is negligible, but if so happens, the entire study (scan) is likely to be redone than manually corrected. Often, QC preview is performed before DICOM rasterization, and therefore all adjustments are simply incorporated into the DICOM conversion transform. As a result, in ever-busy clinical practices no one would really try to manually realign patient image with additional 1-degree rotation: radiologists are trained to view data as is, visually compensating for any deviations, and often using them as additional diagnostic cues.

Finally, archived images are distributed to radiologists' workstations for readings (diagnostic analysis), and this is by far the most important part in the medical image lifecycle. When the images are viewed, they are:
1. Automatically zoomed to fit certain hanging protocols (view layouts)
2. Manually zoomed by the reading radiologists to magnify ROIs (regions of interest)
3. Scrolled ("Cine" tool on most PACS workstations)
4. Occasionally flipped and rotated by multiples of 90 degrees (which does not need interpolation, and can be achieved with the basic symmetries of the image matrix)

Hardly ever are the images rotated by arbitrary angles – as we already mentioned, any position deviations are treated more as additional diagnostic hints, rather than offenses to positional purity. In fact, even flips and 90-degree rotations are left to QC and protocols – any later and arbitrary changes in the original image orientation can have disastrous consequences (think about confusing left and right parts of the brain). As a result, we are really dealing with zooms only – changing image scale by a constant factor (same for $x$ and $y$ coordinates, to preserve image aspect ratio).

However, these zooms become the true bread and butter of diagnostic viewing:
1. Zooms have to be lightning fast. Typically, once ROI is zoomed in, radiologist starts scrolling ("cineing") the images, going through an image series in $z$ direction[5]. An image series can have thousands of images (consider thin slice CTs), and even if interpolating one image takes a fraction of a second, scrolling through a large zoomed series may amount to minutes. This would be completely impractical and unacceptable. Therefore most PACS workstations resort to disabling the interpolation when scrolling through large series, to maintain the fastest image display ("interactive mode"). Then, when the user stops scrolling (releases mouse or scroll button), the interpolation comes back, redrawing the current image with the highest quality ("quality mode"). Interactive interpolation disabling works to a certain extent, but creates an unpleasant effect of jumpy image quality.
2. Zooms are done on certain scales only, even when these scales are not shown in the interface. This is natural: computer processors and interfaces are discrete by definition. For instance, when user zooms with a mouse wheel, he zooms in discrete steps, thus incrementing image size by fixed stepwise zoom factor. Certainly, these stepwise increments are small enough to create an illusion of continuous zooming, but still they are discrete, corresponding to an implicitly defined set of discrete zoom factors.
3. Interpolation quality is everything. Consider zooming an average MR image (256x256 pixels) to a rather small (by radiology display standards) one-megapixel display (1024x1024 pixels). This means replacing each original MR pixel by 4x4=16 simulated pixels, produced by the interpolation algorithm. A few radiologists realize that what they see has so little to do with the original pixels.

---

[4] Picture Archiving and Communication System
[5] If we are dealing with 2D data, such as CR, then high image resolution requires fast zooming as well.

All this brings us to an important conclusion: in diagnostic radiology workflow, image interpolation quality and speed on selected discrete scales are far more important than truly continuous interpolating. In particular, interpolation kernel $h(x)$ continuity is hardly needed or required, but reducing interpolation error $E(h)$ has direct impact on the image diagnostic value. Moreover, using continuous kernels when only stepwise discrete interpolations are performed is highly computationally inefficient. Discrete nature of zooming suggests discrete (stepwise) $h(x)$, which we define as *Interpolation Look-up Tables* (ILUT). Can one build ILUT to meet (Eq.C1) and (Eq.C2)? (Eq.C1) can be easily met. Consider (Eq.C2) and $x=r/Q$, where $r$ and $Q$ are integers (let's call such rational numbers with denominator $Q$ "*Q-rational*"). For Q-rational $x$ its integer shifts ($k+x$) are Q-rational as well, and all interpolation kernel criteria in (Eq.C2) are still satisfied. Hence we define *Q-rational ILUT kernel* $h^Q(x)$ as:

$$h^Q(r) = h(\frac{r}{Q}) \quad \text{(Eq.R)}$$

and the values of $h^Q(x)$ are defined by the look-up table $\{h(x_Q)\}$, where $x_Q = \frac{r}{Q}$, $r=0, 1, 2, …, QL$.

Theorem 2: *Q-rational kernel $h^Q(x)$ interpolation coincides with the continuous interpolation $h(x)$ for Q-rational zoom factors $f=m/Q$.*

**Proof**

The proof immediately follows from the fact that Q-rational zooms of the original integer pixel coordinates $(k,n)$ produce Q-rational coordinates, corresponding to $h^Q(r)$ points. Indeed, to compute convolution in (Eq.I) with discreet $h^Q(r)$, we rewrite

$$J(\frac{r}{Q},\frac{v}{Q}) = \sum_{k=0}^{L} h^Q(r-kQ) \left[ \sum_{n=0}^{L} I(k,n) \times h^Q(v-nQ) \right] \quad \text{(Eq.LUT1)}$$

For any Q-rational zoom factor $f=m/Q$, equation (Eq.LUT1) becomes exact, as if it were computed with a regular continuous kernel:

$$J(\frac{km}{Q},\frac{nm}{Q}) = \sum_{k=0}^{L} h^Q(km-kQ) \left[ \sum_{n=0}^{L} I(k,n) \times h^Q(nm-nQ) \right] \quad \text{(Eq.LUT2)}$$

∎

The advantage of discrete interpolation kernels is obvious – they can be stored with look-up tables, eliminating any need to compute $h(x)$. *Therefore, while gaining speed, we do not lose any quality, as long as we zoom with Q-rational zooms.* What $Q$ should one choose? In our numerical implementation we used $Q=100$, which essentially corresponds to 1% zoom increments - visually perceived as entirely continuous.

The only "expense" of ILUT implementation is storing ILUT values in computer memory. For for typical support size $L=3$, and $Q=100$, we are talking about $QL=300$ double-precision[6] numbers, which is a negligible memory overhead (especially when compared to an average digital image size).

## II. From $L_2$-optimal to ILUT

Our $L_2$-optimal kernels $H_k(x)$ have one obvious drawback – they are not easy to compute. The introduction of ILUT completely eliminates this problem: for a little price of storing $QL$ look-up numbers, we can populate ILUTs with the most optimal values at no computational cost. This is precisely what we wanted to achieve for medical image interpolation.

We tested $L_2$-optimal kernels practically, by subjecting different medical images to the following interpolations: continuous Keys ($L=2$), our discrete optimal $H_2$ ($L=2$, implemented as ILUT for $Q=100$), and continuous Cubic6 ($L=3$). The images were presented to 4 trained radiologists in a blind test, to find out the best-interpolated. Based on visual grading it was determined that our $H_2$ outperforms Keys and provides quality identical to Cubic6, or sometimes better. A few image samples are provided on Figure 4, Figure 5, Figure 6, and Figure 7:

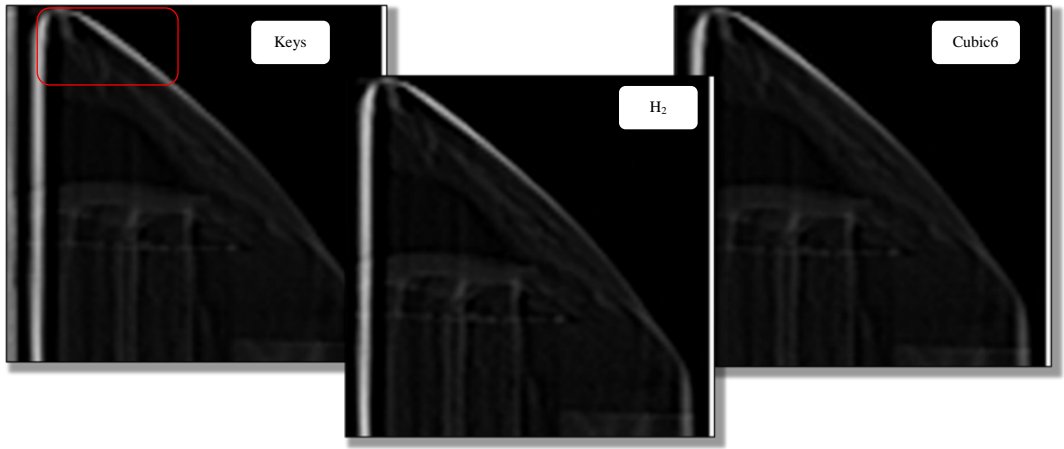

**Figure 4: Interpolation comparison for CT image fragment. Note the strong staircasing artifacts in Keys interpolation, improved with Cubic6 and $H_2$.**

---

[6] It is not hard to reduce ILUT to the integer numbers only, still ensuring sufficient precision in the image interpolation.

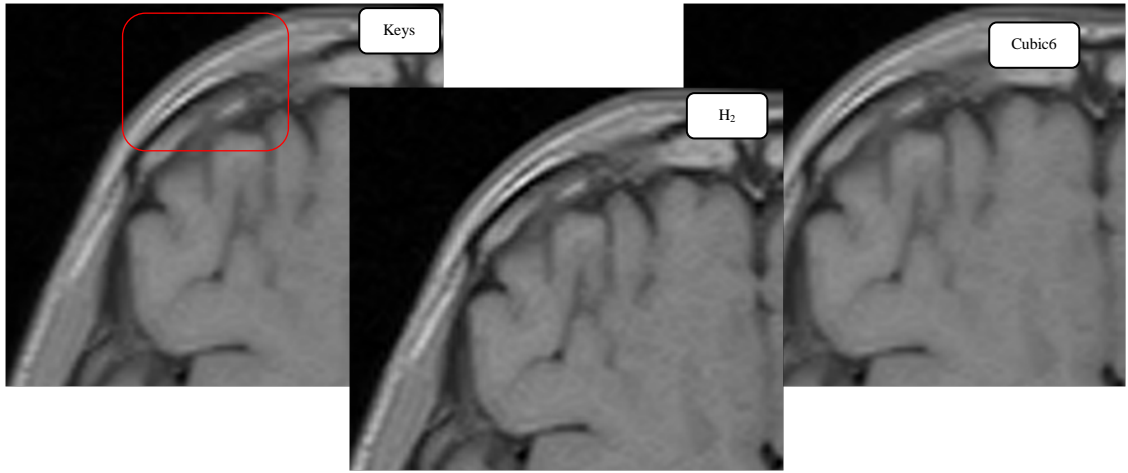

**Figure 5:** Interpolation comparison for MR image fragment. Keys interpolation has visibly lower contrast and more artifacts on the edges, and Cubic6 has the same problem. Optimal $H_2$ significantly reduces the artifacts.

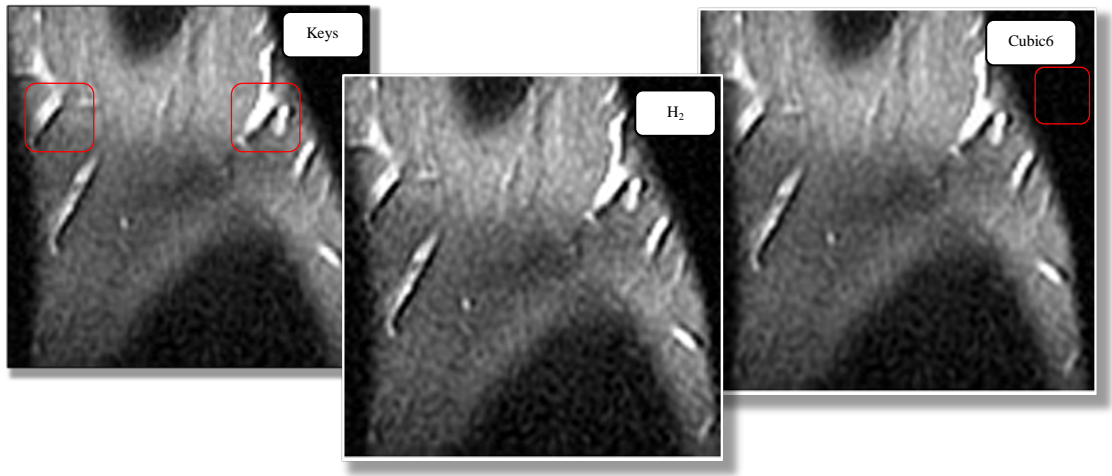

**Figure 6:** Interpolation comparison for MR image fragment. Keys interpolation has visibly lower contrast and more artifacts on the edges. This is improved with Cubic6 and $H_2$, but note that Cubic6 visibly smoothes image details, making some areas (as selected background fragment) nearly uniform. $H_2$ image seems to be the sharpest of the three.

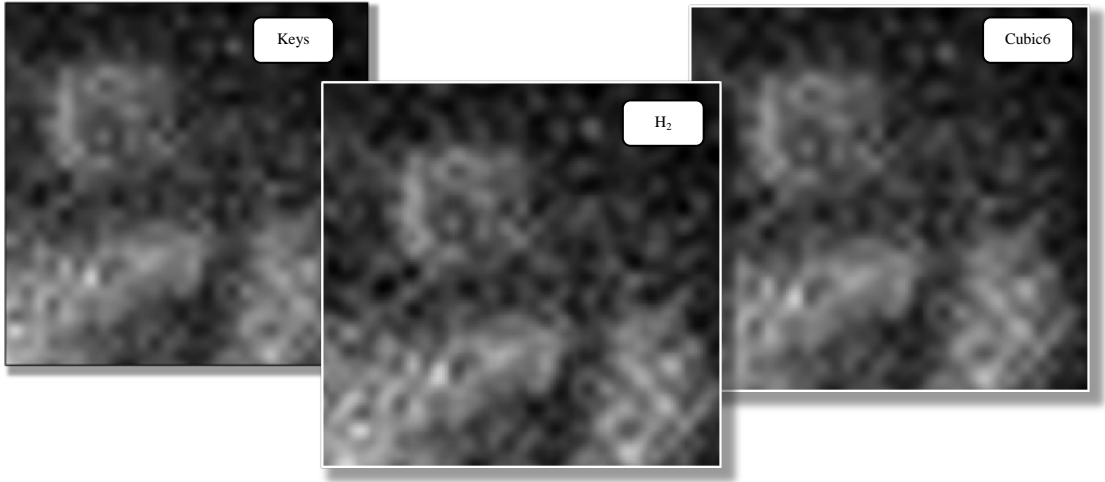

**Figure 7: Interpolation comparison for nuclear medicine image fragment. Keys interpolation suffers from visible loss of contrast, improved in Cubic6 and $H_2$.**

It needs to be mentioned that $F(H_2)(t)$ frequency maximum around $t=0.2$ (Figure 3, center) plays an interesting role: it "magnifies" local frequencies. It appeared in our tests that this magnification is perceived visually as an enhancement rather than a distortion. This proves that the monotonicity of Fourier transforms of the interpolation kernels is hardly needed, and can be visually outperformed by our $L_2$-optimal kernels.

### III. FINAL CONSIDERATIONS

As we already explained, digital image rotations are very rare in the routine radiological workflow, but we do not want to rule them out completely. Rotation by angle α is equivalent to multiplying 2D pixel coordinates by the rotation matrix

$$M = \begin{bmatrix} \cos(\alpha) & \sin(\alpha) \\ -\sin(\alpha) & \cos(\alpha) \end{bmatrix}$$

Generally, cos/sin are not $Q$-rational for any $Q$, but we can approximate them closely with $Q$-rational $c_Q=n/Q$ and $s_Q=m/Q$, transforming the coordinate grid by

$$M_Q = \begin{bmatrix} n/Q & m/Q \\ -m/Q & n/Q \end{bmatrix}$$

This will deviate from the original angle α by dα=(a-atan(m/n)), and introduce additional zoom of $\sqrt{n^2+m^2}/Q$. However, both deviations can be minimized with sufficiently large $Q$, producing visibly-identical result.

Oscillating pattern in $F(H_3)$ (Figure 3, right) can be viewed as another concern, as it can be associated with ringing artifacts in the interpolated images. From our experience, the presence of ringing may or may not be visible, depending on the nature of the images. Nonetheless, to make this study complete, we can suggest two ways to modify $H_k(x)$ behavior:

1. Averaged kernels, such as $A_3(x) = wH_3(x)+(1-w)h_{Cubic6}(x)$, $0<w<1$, still satisfy (Eq.C1) and (Eq.C2), and provide a simple yet efficient way to alleviate certain $H_k(x)$-specific artifacts by blending them with more smoothing polynomial kernels.
2. Adding additional constraints to (Eq.var), in the form of Lagrangian multipliers, provides an infinite range of possibilities for changing certain $H_k(x)$ properties – at the expense of higher FAE.

We performed initial experiments in both directions, but since they depend on the additional assumptions, would like to leave them outside the scope of this work.

In conclusion, we would like to summarize the use of $L_2$-optimal $H_k(x)$, reduced to discrete ILUT, as an ideal solution for the routine medical imaging. This approach combines high computational efficiency with superior image quality, which makes it particularly practical in digital radiology.